# AsPOS: Assamese Part of Speech Tagger using Deep Learning Approach

Dhrubajyoti Pathak*, Sukumar Nandi† and Priyankoo Sarmah‡
*Centre for Linguistic Science and Technology*
*Indian Institute of Technology Guwahati*
North Guwahati, India
Email: *drbj153@iitg.ac.in, †sukumar@iitg.ac.in ‡priyankoo@iitg.ac.in

*Abstract*—Part of Speech (POS) tagging is crucial to Natural Language Processing (NLP). It is a well-studied topic in several resource-rich languages. However, the development of computational linguistic resources is still in its infancy despite the existence of numerous languages that are historically and literary rich. Assamese, an Indian scheduled language, spoken by more than 25 million people, falls under this category. In this paper, we present a Deep Learning (DL)-based POS tagger for Assamese. The development process is divided into two stages. In the first phase, several pre-trained word embeddings are employed to train several tagging models. This allows us to evaluate the performance of the word embeddings in the POS tagging task. The top-performing model from the first phase is employed to annotate another set of new sentences. In the second phase, the model is trained further using the fresh dataset. Finally, we attain a tagging accuracy of 86.52% in F1 score. The model may serve as a baseline for further study on DL-based Assamese POS tagging.

## I. INTRODUCTION

Natural language processing (NLP) is an automatic approach to analyze texts or speech using different technologies with the help of a machine (computer). It has become part and parcel of daily life. It is also described as a computerized approach to process, understand and generate natural language. Thus it enhances human-to-human conversation and enables human-to-machine conversation through effective processing of texts or speeches. Effective preprocessing of natural language text is essential for various natural language processing (NLP) tasks. Part of speech (POS) tagging is one of the most important areas and building block and application in natural language processing. POS tags are shown to be useful features in various NLP tasks, such as named entity recognition (NER) [1], machine translation [2], Question Answering [3] and constituency parsing [4]. Therefore, a well-performing POS tagger plays a crucial role in developing a successful NLP system for any language. POS tagging is the process of automatic assignment of grammatical class types to words in a sentence such as noun, adjective, pronoun, verb, adverb, conjunction etc. Various approaches have been deployed for automatic POS tagging. These approaches can broadly be categorized into two.

1) Traditional approaches
   - Rule-based approaches, which do not require annotated corpus as they are mainly based on manually hand-crafted rules to assign tags to words in a sentence.; for instance, a word that follows adjectives in a sentence is tagged as Noun. Rules can be compiled from the linguistic features of the language, such as syntactic, lexical, and morphological information [5].
   - Unsupervised learning approaches, which are mainly based on unsupervised algorithms without hand-crafted labeled training data; The typical approach in unsupervised learning is clustering [6]. Basically, the techniques is based on lexical resources (e.g. WordNet), lexical patterns, and on statistics that are computed on a large unannotated corpus.
   - Feature-based supervised learning approaches, which are mainly based on supervised learning or machine learning algorithms with careful feature engineering; Machine learning algorithms such as Decision Trees [7], Maximum Entropy Models [8], Hidden Markov Models (HMM) [9], Support Vector Machines (SVM) [10], and Conditional Random Fields (CRF) [11] are applied to learn a model to recognize similar patterns from unseen data.

2) Deep learning (DL) based approach
   - DL-based approaches, which automatically extract data representations required for the classification, identification, or detection from raw input in an end-to-end manner. It is a data-intensive approach that come with a better result than the traditional methods ( HMM, SVM, Clustering, etc.). It requires a large annotated dataset to train a sequence labeling task. Common sequential deep learning method such as MLP [12], GRU [13], CNN, RNN [14], LSTM [15], BiLSTM [16] and Transformer [17] are applied automatically extract latent representations or linguistic characteristic that required for classification or detection task.

The traditional feature-based or rule-based approaches require a considerable amount of engineering skill and language expertise. Developing a well-performed rule-based POS tagging system is challenging for morphologically rich high inflectional languages due to complex grammatical rules and attests to a large vocabulary. Another issue to address in Part of speech tagging is the language ambiguity, where the meaning (or class type) of a word changes with the context of a sentence. Therefore, sometimes it becomes challenging to identify the correct tag of a word appearing in a sentence.

Recent advances in DL-based POS models achieve state-of-the-art (SOTA) accuracy and become more dominant. DL-based POS models are shown to be effective in automatically learning useful representations and underlying linguistics characteristics from text. Several well-performing POS taggers exist for resource-rich languages using the word embedding and deep learning neural network [18]. However, it requires basic resources such as a large amount of annotated training data, a well-trained word embeddings etc. [19] to build well-performing models in DL-based approach. Hence the neural models are not well experimented on resource-poor languages, which have limited resources of annotated training data.

Assamese or Asamiya (ɔxɔmiyɑ) is an Indo-Aryan language spoken mainly in Assam, a state of northeast India. It is a descendant of Magadhi Prakrit and bears affinities with Bengali, Hindi, and Odia. Modern Assamese uses the Assamese script, which is developed from the Brahmi script. Assamese script is similar to Bengali script except for two characters, where Assamese differs from Bengali in one letter (ৰ) for the /r/ sound, and an extra letter (ৱ) for the /w/ or /v/ sound. Assamese is a morphologically rich, highly inflectional, and relatively free-order language.

Although it is one of the scheduled languages of India, spoken by more than 25 million people. However, Assamese is also a low-resource language, which does not have quality resources to train a well-performed POS tagger. Although there are few works on Assamese POS tagging; however, we do not find any study about DL-based POS taggers for Assamese language. The existing taggers are either rule-based or supervised learning-based. These motivate us to take the challenge to develop a deep learning-based approach for Assamese POS tagging. We choose the BiLSTM-CRF architecture to build a deep neural-based POS model for Assamese. The BiLSTM-CRF framework [20] shows high-performance accuracy in sequence tagging on rich-resource languages. The accuracy of the standard POS tagging model for English is 98.19% [21].

In this paper, our contributions can be summarized as follows:
- We preprocess the annotated POS corpus obtained from TDIL [22] and prepare it for training.
- We try out eleven different pre-trained word embeddings in the POS tagging task, including Glove, Word2Vec, Fasttext, BytePair, Character embedding, BERT, ELMO, FlairEmbeddings, MuRIL, XLM-R, and IndicBERT.
- Using the best-performing tagging model, we prepare a new POS-tagged corpus and then manually improve the corpus for future training.
- The top-performing model is further trained and achieves an F1 score of 86.52% accuracy.
- The dataset and POS tagging model[1] are made accessible online for the research community.

In our literature study, we find some previous works on Assamese POS tagging; [23] developed a POS tagger for the Assamese language using the HMM and Viterbi algorithms with an accuracy of F1 score 85.64%. [24] presented a semi-automated tagger that is based on rules and a database. [25] proposed a POS tagger using Conditional Random Field (CRF) and Transformation Based Learning (TBL) with an accuracy of 87.17% and 67.73% respectively. [26] present a POS system for annotation of English-Assamese code-mixed text. The accuracy of the tagger stands at 84.00%. However, in the experiment result, the existing works are not compared because-

- As we could not find any of the existing paper's code or model in the public domain to try it practically.
- The existing taggers accuracy are measured not on F1 score but the fraction of correctly tagged words to the total number of total words in the test set.
- The taggers are not complete i.e. a) they are limited to simple sentences b) Their tag set is not based on BIS c) they are tested in a small set limited to a domain.
- They are not based on DL method.

Considering this, the aim of this paper is to report the effectiveness of a DL-based POS tagger for Assamese and the rest of the paper is organized as follows. A brief description of the development of the DL-based tagger along with the details of the dataset and pre-trained word embeddings are reported in Section II. The details of the experimental setup are presented in Section III. The results and analysis are reported in Section IV. Section V provides the error analysis from the results and the paper is concluded in Section VI.

## II. Deep neural based tagger

There are various stages in building a DL-based POS tagging model. It can mainly be divided into three- i) Collection/preparation of properly annotated dataset ii) Embedding the words in sentences using a well trained, appropriate word embedding iii) Training the model for tagging. In the following sections, we explain these stages elaborately.

### A. Annotated Dataset

In nature, DL algorithms are resource hungry in terms of computational resources. It requires a large as well as properly labeled dataset to make POS tagging model efficient in identifying the label of unknown words. However, preparing a suitable corpus for a language is tedious and time-consuming. It needs plenty of language resources and requires language experts to verify the annotated dataset. This is a challenging task for a low resource language such as Assamese. In our literature study, we find only one publicly available POS annotated dataset. The annotated dataset [22] was tagged by language experts manually as part of Indian Languages Corpora Initiative (ILCI), Govt. of India. We acquired that for the training and evaluation. The BIS tagged datasets consist of original Assamese texts from newspapers, magazines, etc., over multiple domains, such as agriculture, art & culture, economy, entertainment, etc., with 35K sentences and 404K words.

---

[1]https://huggingface.co/dpathak/aspos_assamese_pos_tagger

We prepare the labeled dataset in column format in which each line represents a word, and each column represents the POS tag of the corresponding word. An empty line separates sentences in the dataset. In table I, the sample format is shown.

TABLE I
Dataset format

| Word | Tag |
|---|---|
| দিল্লী | N_NNP |
| ভাৰতৰ | N_NNP |
| ৰাজধানী | N_ANN |
| । | RD_PUNC |
|  |  |
| প্ৰধানমন্ত্ৰী | N_ANN |
| দিল্লীত | N_NNP |
| থাকে | V_VAUX |
| । | RD_PUNC |

The first column of the dataset is the word itself and the second one is the PoS tag.

*1) Tagset used:* Bureau of Indian Standards (BIS) tagset has been declared the national standard for annotating Indian language data. So, we used the BIS tagset, which contains 41 tags with eleven (11) top-level categories. The description of all tagsets is presented in Table II.

### B. Pre-trained word embeddings

There are various pre-trained word embeddings (language models) available for Assamese language. These embeddings are mostly trained using Wikipedia, which comprises approximately 8 million tokens. Comparatively, the corpus size is one of the smallest among all the Indian languages. In our study, we find that these pre-trained word embeddings are not well studied in downstream tasks. Therefore, we try to cover all the Assamese pre-trained word embedding to train our POS tagging model in our experiment. These includes Word2Vec [27], Glove [28], Byte-Pair Embeddings [29], FastText embedding [30], BERT [31], ELMo [32], XLM-R [33], MuRIL [34], IndicBert [35], FlairEmbeddings [36]. We also cover the Character Embeddings [37]. In the following section, we describe the training.

### C. Model training

In this section, we present the details about the training of the POS tagging model. We use conditional random field (CRF), Bidirectional long short-term memory (BiL-STM) with CRF, and GRU with CRF architecture to train the POS model using Flair framework [21]. However, we observe that the combination of BiLSTM with CRF performs better. Therefore, in our experiments, we use only BiLSTM-CRF [20] sequence tagging model with different variants of word-embedding setup.

*1) Training details:* Different hyperparameters are explored in several configurations during training. In Table III, the hyperparameters are mentioned. We use the same set of hyperparameters in all our experiments. To account for memory constraints, we use a fixed mini-batch size of 16. We use the early stopping technique if there is no improvement in the validation data accuracy. Learning Rate Annealing factor is also used for early stopping.

**Training time**- We conducted all our experiments on Nvidia Tesla P100 GPU (3584 Cuda Cores). It takes an average of twenty hours to complete training and test. Although the max epoch is kept at 100, in most of cases, the training completes before the max epoch due to the early stopping technique.

## III. Experimental Setup

In this section, we discuss the development of the POS tagging model. The development process can be divided into two phases. In the first phase, we use the acquired dataset for training in all the experiments. In the second phase, the top-performing model is used as a tagger to tag new sentences. After that, the new tagged sentences are manually corrected and used for further training.

### A. Phase I

All our experiments for the development of POS tagging model are run on different language models discussed in II-B. We use the Huggingface Transformers library [38] for XLM-R, IndicBERT, MuRIL, Github library - Glove[2], Byte-Pair Embedding[3], Flair Embedding[4], Fassttext library [5] and Word2vec[6] to set up our experiments. There are two rounds of experiments conducted on the TDIL annotated dataset. In the first round, each pre-trained word embedding model is employed separately for training. The training on the same dataset using each embedding separately provides us with the performance comparison of these embeddings on the POS tagging task.

[2] https://github.com/stanfordnlp/GloVe
[3] https://github.com/bheinzerling/bpemb
[4] https://github.com/flairNLP/flair/blob/master/resources/docs/embeddings/FLAIR_EMBEDDINGS.md
[5] https://fasttext.cc/docs/en/pretrained-vectors.html
[6] https://www.cfilt.iitb.ac.in/~diptesh/embeddings/

TABLE II
Details of tagset

| S.No | Category | Type | Tag |
|---|---|---|---|
| 1 | Noun | Proper Noun | N_NNP |
| | | Common Noun | N_CNN |
| | | Verbal Noun | N_VNN |
| | | Abstract Noun | N_ANN |
| | | Material Noun | N_MNN |
| | | Noun (Location) | N_NST |
| | | Noun (unclassified) | N_NN |
| 2 | Pronoun | Personal | PR_PRP |
| | | Reflexive | PR_PRF |
| | | Reciprocal | PR_PRC |
| | | Relative | PR_PRL |
| | | Wh-words | PR_PRQ |
| | | Indefinite | PR_PRI |
| 3 | Demonstrative | Deictic | DM_DMD |
| | | Relative | DM_DMR |
| | | Wh-words | DM_DMQ |
| | | Indefinite | DM_DMI |
| 4 | Verb | Auxiliary Verb | V_VAUX |
| | | Main Verb | V_VM |
| | | Transitive | V_VBT |
| | | In-transitive | V_VBI |
| 5 | Adjective | Proper Adjective | J_PJJ |
| | | Verbal Adjective | J_VJJ |
| | | Adjectival Adverb | J_JJ |
| 6 | Adverb | | RB |
| 7 | Post Position | | PSP |
| 8 | Conjunction | Conjunction | CC_CCD |
| | | Co-ordinator | CC_CCS |
| 9 | Particles | Particles (unclassified) | SUF |
| | | Classifier | RP_RPD |
| | | Interjection | RP_INJ |
| | | Negation | RP_NEG |
| | | Intensifier | RP_INTF |
| 10 | Quantifiers | General | QT_QTF |
| | | Cardinals | QT_QTC |
| | | Ordinals | QT_QTO |
| 11 | Residuals | Foreign word | RD_RDF |
| | | Symbol | RD_SYM |
| | | Punctuation | RD_PUNC |
| | | Echowords | RD_ECH |
| | | Unknown | RD_UNK |

TABLE III
Fine-tuning hyper-parameter details used in training

| Hidden layer size | Hidden layer | Word dropout | Initial Learning rate | Max epochs | Sequence length |
|---|---|---|---|---|---|
| 512 | 2 | 0.05 | 0.01 | 100 | 128 |

In the second round, we use the stacked method to train the POS tagging model further. By the stacked method, we get to know the combining performance of one embedding with other embeddings during training. It has been shown that stacked embeddings give state-of-the-art accuracy across many sequence labeling tasks [21]. We consider the top two performing models from the first round for further training in the stacked method.

*B. Phase-II*

In Phase-II, we use the best performing tagging model from Phase-I, which achieves the highest accuracy for further training. The tagging model is used for the annotation of new sentences. Thereafter, the newly annotated sentences are manually corrected by following Assamese grammar [39, 40, 41, 42] and dictionary [43, 44]. The new dataset comprises 42k tokens and is used for further training.

In this phase, we use the same configuration with the combination of the word embeddings of the best-performing model.

IV. Result and Analysis

In this section, we discuss the outcomes of the two phases of the experiment. During the literature review, we found no studies on the performance of any Assamese pre-trained word embeddings on sequence tagging tasks; therefore, we attempted to cover all recent state-of-the-art word embeddings to analyze their performance in the POS tagging task.

TABLE IV
PHASE-I: SUMMARY OF THE PERFORMANCE OF INDIVIDUAL WORD-EMBEDDINGS IN POS TAGGING TASK

| Embeddings | F1 score |
|---|---|
| WordEmbeddings (Glove) | 48.6% |
| Word2Vec | 39.54% |
| FastTextEmbeddings | 69.81% |
| BytePairEmbeddings | 70.99% |
| CharacterEmbeddings | 46.03% |
| BERT Embedding | 70.33% |
| ELMO | 71.32% |
| FlairEmbeddings (Multi) | 70.76% |
| MuRIL | 72.86% |
| XLM-R | 70.01% |
| IndicBert | 72.93% |

TABLE V
PHASE-I: SUMMARY OF THE PERFORMANCE OF STACKED WORD-EMBEDDINGS IN POS TAGGING

| Stacked Embeddings | F1 score |
|---|---|
| MuRIL + Glove | 74.31% |
| MuRIL + Word2Vec | 74.46% |
| MuRIL + FastTextEmbeddings | 73.7% |
| MuRIL + BytePairEmbeddings (BPEmb) | 74.14% |
| MuRIL + CharacterEmbeddings | 73.84% |
| MuRIL + BERT | 73.95% |
| MuRIL + ELMO | 73.16% |
| MuRIL + XLM-R | 70.42% |
| MuRIL + IndicBert | 74.25% |
| **MuRIL + FlairEmbeddings** | **74.62%** |
| IndicBert + Glove | 74.06% |
| IndicBert + Word2Vec | 73.94% |
| IndicBert + FastTextEmbeddings | 72.89% |
| IndicBert + BytePairEmbeddings (BPEmb) | 73.39% |
| IndicBert + CharacterEmbeddings | 73.25% |
| IndicBert + BERT | 72.66% |
| IndicBert + ELMO | 73.28% |
| IndicBert + XLM-R | 72.98% |
| **IndicBert + FlairEmbeddings** | **74.21%** |

TABLE VI
PHASE-II: POS TAGGING PERFORMANCE OF THE FINAL MODEL (MURIL + FLAIREMBEDDINGS) USING DIFFERENT TAGGING TECHNIQUE

| Embeddings | Tagging model | F1 score |
|---|---|---|
| MuRIL + FlairEmbeddings | CRF | 84.88% |
|  | **BiLSTM + CRF** | **86.52%** |

**Phase I**: In the first round, we observe that models with contextual embeddings, such as BERT, IndicBERT, XLM-R, MuRIL, ELMo, and FlairEmbeddings, as well as sub-word embeddings, such as Fasttext and BytePair, achieve similar performance in POS tagging, whereas word embeddings, such as Glove, Word2Vec, and Character Embeddings, perform poorly in tagging task. The results of our first round of experiments are reported in Table IV.

In the second round, we pick the configuration of the top two performers, i.e., the model with MuRIL word embedding and the other with IndicBERT word embedding. The F1 score performance accuracy of both models is 72.86% and 72.93%, respectively. In this round, we examine the performance of MuRIL and IndicBert in combination with the remaining word embeddings. The results of the second round of the experiment are shown in Table V. We see that the stacked method enhances the tagging performance in all the configurations. When MuRIL is paired with FlairEmbedding, the maximum tagging accuracy is attained with an F1 score of 74.62%.

**Phase II**: In Phase II, we use the best-performing model (MuRIL+FlairEmbedding) from Phase I to annotate a new set of sentences from another corpus. The new corpus is comprised of the translated text of the speech delivered by the Indian Prime Minister and published by the Press Information Bureau, Government of India[7]. Following the completion of the manual editing of the annotated text, the newly created dataset was utilized for additional training.

In Phase II, we set up two configurations employing two sequence tagging models: CRF and BiLSTM. With the two final settings in hand, we used an evaluation-based technique to find the best model for POS tagging. Table VI summarizes the findings of the experimental results of the tagging performance. The F1 score for the BiLSTM-CRF model is 86.52. Figure 1 depicts the learning curve of the training and validation scores. Convergence of the learning curve suggests that the train and validation datasets are reasonably representative of POS tagging.

---
[7] https://pib.gov.in/indexd.aspx/

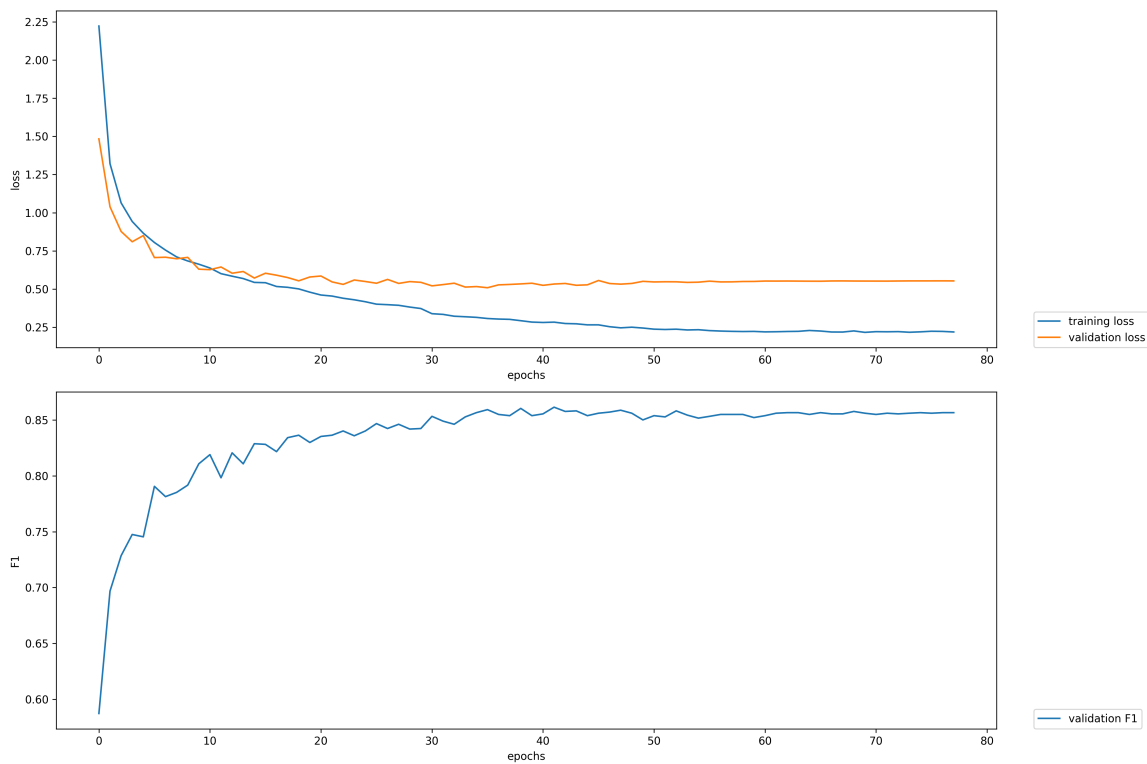

Fig. 1. Learning curve of top performing model- MuRIL with Flair Embedding

As native speakers, we see inconsistencies in the already annotated dataset, such as comma (,) being categorized as both a Punctuation mark (PUNC) and a Symbol (SYM) in some occurrences. In order to evaluate the performance of the model on the newly annotated dataset, we use solely the new dataset to train/test a new model with the same architectures (MuRIL+FlairEmbeddings embeddings in BILSTM-CRF). In addition, we train the same model with the existing+new dataset while keeping the evaluation set the same, resulting in a ≈3% improvement in performance.

## V. Error Analysis

We perform an error analysis of the test result of our final model. We describe the error details as follows-

- The model wrongly predicts a tag when a class type of a noun word changes to an adjective. It happens when it describes another noun, e.g., *'Assamese language'*; although *Assamese* is a noun, in this particular case, it is being used to describe another noun word - *language*. Therefore its class changes to an adjective.
- Even though Assamese is an SOV word order language, it allows parsing. Hence, due to the change of context because of parsing, it becomes difficult to figure out the correct tag – JJ or N_NN, V_VM or N_NN, and RB or N_NN. Therefore, sometimes the taggers incorrectly tag as N_NN for JJ, V_VM, and RB.
- Unlike English, Assamese orthography does not indicate proper nouns in the written form. Apart from that, several common nouns may also be used as proper nouns [45]. Therefore, it is difficult to automatically differentiate between a proper noun and a common noun. Extracting the proper Noun from a sentence requires an understanding of the statement's context.
- The most common error is confusion with the Main Verb (V_VM) as Auxiliary (V_VAUX) and vice versa.
- Sometimes, the model tags roman numbers as foreign words (RD_RDF), whereas they are Quantifiers (QT_QTF) in the test set.

## VI. Conclusion

In this paper, a Deep Learning-based POS tagger for Assamese was presented. We believe this is the first attempt to develop a DL-based Assamese POS tagger. Additionally, we prepared an annotated dataset. As part of the development process, we also presented an evaluation of the performance of multiple pre-trained Assamese word embeddings for the POS tagging task. The performance comparison assisted us in determining the best configuration for a POS tagging model. Although our POS tagger could not achieve state-of-the-art accuracy as that of resource-rich, we believe our POS tagging model can be a baseline for future research. Our contributions can benefit the research community by leveraging our model, dataset, and word embedding's performance outcomes in sequence tagging tasks, particularly for Assamese and other highly inflectional, morphologically rich, low-resource languages when selecting word-embedding models for downstream tasks.